\documentclass{article}

\PassOptionsToPackage{numbers,compress}{natbib}
\usepackage[dblblindworkshop, final]{neurips_2026}
\workshoptitle{TS-LIMITS}

\usepackage[utf8]{inputenc} 
\usepackage[T1]{fontenc}    
\usepackage{float}          
\usepackage{hyperref}       
\usepackage{acronym}        
\usepackage{url}            
\usepackage{booktabs}       
\usepackage{graphicx}       
\usepackage{amsmath}        
\usepackage{enumitem}       
\usepackage{amsfonts}       
\usepackage{nicefrac}       
\usepackage{microtype}      
\usepackage{xcolor}         
\usepackage{todonotes}
\usepackage{booktabs}
\usepackage{lipsum}
\usepackage{optidef}
\usepackage{mathtools}
\usepackage{placeins}

\mathtoolsset{showonlyrefs}

\title{Time-series Foundation Models for Predictive Control: The Role of Excitation}

\author{%
  Mazen Amria$^{1}$ \quad
  Jasper Hoffmann$^{1}$ \quad
  Philipp Bordne$^{1}$ \quad
  Anna Rothenhäusler$^{1}$
  \AND
  Lilli Frison$^{1}$ \quad
  Harald Taxt Walnum$^{2}$ \quad
  Sebastien Gros$^{2}$ \quad
  Joschka Bödecker$^{1}$ \\
  \\
  $^{1}$University of Freiburg, Germany \\
  $^{2}$Norwegian University of Science and Technology, Norway
}

\usepackage{acronym}

\newacro{TSFM}{time-series foundation model}
\newacro{MPC}{model predictive control}
\newacro{HVAC}{heating, ventilation, and air conditioning}
\newacro{ICL}{in-context learning}
\newacro{APRBS}{amplitude-modulated pseudo-random binary sequences}
\newacro{EWMA}{exponentially weighted moving average}
\newacro{MAE}{mean absolute error}
\newacro{iCEM}{improved cross-entropy method for trajectory optimization}

\newacro{DM}{decision making}
\newacro{AD}{automated driving}
\newacro{SOC}{discrete-time stochastic optimal control}

\newacro{NLP}{nonlinear program}
\newacro{ODE}{ordinary differential equation}
\newacro{MPC}{model predictive control}
\newacro{MHE}{moving horizon estimation}
\newacro{NMPC}{nonlinear model predictive control}
\newacro{ENMPC}{economic nonlinear model predictive control}
\newacro{MPPI}{model predictive path integral control}
\newacro{SPC}{subspace predictive control}
\newacro{QP}{quadratic program}
\newacro{MIQP}{mixed-integer {\ac{QP}}}
\newacro{MILP}{mixed-integer \ac{LP}}
\newacro{MINLP}{mixed-integer \ac{NLP}}
\newacro{MI}{mixed-integer}
\newacro{MIP}{mixed-integer programming}
\newacro{MPCC}{mathematical program with complementarity constraints}
\acrodefplural{MPCC}{mathematical programs with complementarity constraints}
\newacro{BB}{branch-and-bound}
\newacro{SQP}{sequential quadratic programming}
\newacro{RNN}{recurrent neural network}
\newacro{OCP}{optimal control problem}
\newacro{LQR}{linear quadratic regulator}
\newacro{iLQR}{iterative linear quadratic regulator}
\newacro{SLQ}{sequential linear quadratic programming}
\newacro{DDP}{differential dynamic programming}
\newacro{MCP}{mixed complementarity problem}
\newacro{IP}{interior point}
\newacro{ADMM}{alternating direction method of multipliers}
\newacro{RTI}{real time iteration}
\newacro{POMDP}{partially observable \ac{MDP}}
\newacro{CEM}{cross-entropy method}
\newacro{GGN}{generalized Gauss-Newton}
\newacro{KKT}{Karush-Kuhn-Tucker}
\newacro{IFT}{implicit function theorem}
\newacro{RL}{reinforcement learning}
\newacro{DP}{dynamic programming}
\newacro{LSTM}{long short-term memory}
\newacro{NN}{artificial neural network}
\newacro{PPO}{proximal policy optimization}
\newacro{PG}{policy gradient}
\newacro{DPG}{deterministic \ac{PG}}
\newacro{SPG}{stochastic \ac{PG}}
\newacro{TD}{temporal difference}
\newacro{SAC}{soft actor-critic}
\newacro{PI}{policy iteration}
\newacro{SARSA}{state action reward state action}
\newacro{IL}{imitation learning}
\newacro{MLE}{maximum likelihood estimation}
\newacro{MDP}{Markov decision process}
\newacroplural{MDP}[MDPs]{Markov decision processes}
\newacro{DDPG}{deep deterministic \ac{PG}}
\newacro{DQN}{deep Q-networks}
\newacro{TD3}{twin-delayed actor-critic}
\newacro{VI}{value iteration}
\newacro{TRPO}{trust region policy optimization}
\newacro{BC}{behavior cloning}
\newacro{GAN}{generative adversarial network}
\newacro{GPS}{guided policy search}
\newacro{GP}{Gaussian process}
\newacro{leap-c}{Learning for Predictive Control}
\newacro{FA}{function approximator}
\newacro{ADP}{approximate \ac{DP}}
\newacro{LMPC}{linear \ac{MPC}}

\newacro{FF}{feed forward network}
\newacro{DNN}{deep neural network}
\newacro{OMD}{optimal model design}

\newacro{UGV}{unmanned ground vehicle}

\newacro{RWE}{real-world experiments}

\definecolor{saczopcolor}{HTML}{EE8636}
\definecolor{sacfopcolor}{HTML}{8E69B8}

\colorlet{saczopdark}{saczopcolor!85!black}
\colorlet{sacfopdark}{sacfopcolor!85!black}

\begin{document}

\maketitle

\begingroup
\renewcommand{\thefootnote}{}
\footnotetext{Correspondence to: Mazen Amria <amriam@cs.uni-freiburg.de>}
\addtocounter{footnote}{-1}
\endgroup

\begin{abstract}
Deploying model predictive control (MPC) requires constructing or identifying a predictive model for each target system. Time-series foundation models (TSFMs) offer an attractive option thanks to strong zero-shot forecasting capabilities across systems. However, low forecast error does not guarantee that a TSFM captures the system's response to the alternative actions considered by the controller. We study this gap using residential heat-pump control as a test bed, measuring the agreement between predicted and ground-truth effects of control interventions. Importantly, we find that TSFMs can recover the system’s input–response relationship when the context contains sufficient independent control excitation. Common fine-tuning pipelines and feature smoothing reduce, but do not eliminate, the need for in-context excitation. Our results indicate that current TSFMs used for predictive control require sufficiently informative control variation in the inference context. Initial closed-loop results show promise for shorter context windows.
\end{abstract}

\section{Introduction}
\Acp{TSFM} have emerged as strong zero-shot forecasters, with impressive transfer across datasets and domains~\cite{ansari2024chronos}. This makes them promising dynamic models for \ac{MPC}, potentially avoiding the need to construct or identify a model specifically for each target system. However, forecasting accuracy alone is not sufficient for control. Standard forecasting evaluates predictions on trajectories generated by a fixed controller that produced the training data, whereas MPC queries the model under candidate control inputs that may differ substantially from those observed. A model may therefore achieve low forecasting error by capturing passive dynamics, exogenous disturbances, or regularities of the data-generating policy, without correctly modeling the effect of out-of-distribution control inputs. Effective planning instead requires separating the effect of a control input from these of other sources of variation.

\begin{figure*}[!t]
    \centering
    \includegraphics[width=\textwidth]{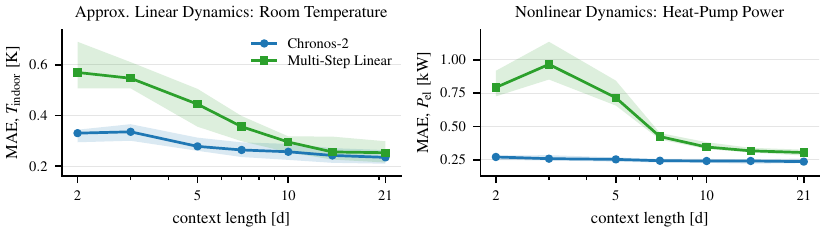}
    \caption{\textbf{Strong zero-shot forecasting on real-world building data.}
    Average \ac{MAE} of Chronos-2~\citep{chronos2} and a multi-step linear predictor for approximately linear room-temperature dynamics and nonlinear heat-pump power dynamics across context lengths. See \autoref{app:real_probe} for details.}
    \label{fig:real_world_forecasting}
\end{figure*}

Replacing system-specific identification with a transferable pretrained model is particularly attractive in building energy management, where predictive control can require substantial effort to construct and identify system-specific models. Building thermal dynamics vary across systems and operating conditions, are strongly influenced by disturbances such as weather and occupancy, and are observed through heterogeneous sensing, actuation, and control hardware.
Thus, such systems are intrinsically stochastic and only partially observed.
In this setting, data-driven linear multi-step predictors can be effectively integrated into \ac{MPC} frameworks \cite{reinhardt2025data}.
On real-world building data, however, modern \acp{TSFM} can achieve substantially better nominal forecasting performance.
As shown in \autoref{fig:real_world_forecasting}, zero-shot Chronos-2~\citep{chronos2} outperforms a multi-step linear predictor for both room-temperature and heat-pump power forecasting across context lengths.

Yet this forecasting advantage does not imply that the model captures the input-output relationships required for control. Previous work has raised concerns about this distinction. At the same time, recent \acp{TSFM} such as Chronos-2~\citep{chronos2} provide richer conditioning mechanisms, including known future covariates and multivariate forecasting. These developments motivate a more systematic evaluation of recent \acp{TSFM}'s ability to capture control-relevant input-response behavior rather than merely predicting trajectories accurately under the data-generating distribution.

We investigate this distinction in residential heat-pump control, using it as a controlled test bed for separating forecasting performance from control-relevant dynamics. We construct matched intervention probes in which the initial state and disturbances are held fixed while the future supply-temperature trajectory is varied, allowing us to assess nominal forecast accuracy independently from input-response fidelity. We further study whether input-response fidelity can be improved through fine-tuning on independently excited trajectories, excitation in the inference context, and multiscale representations of the control input.

Our contributions are threefold: (i) we find that, in simplified synthetic experiments and under certain conditions, \acp{TSFM} can identify the control-relevant input-output relationship; (ii) we show that fine-tuning on closed-loop operational data improves predictions of the response to alternative controls, with greater improvements when fine-tuning on excited data, while leaving nominal forecasting accuracy along the factual control trajectory largely unchanged; and (iii) we show that excitation in the inference context is a main mechanism enabling system identification, while fine-tuning provides a complementary prior that reduces the amount of informative excited context required when transferring to unseen buildings, but does not eliminate the need for excitation.

\section{Time-series Foundation Models for Model Predictive Control}\label{sec:tsfm_mpc}

In the following, we further discuss the potential of \acp{TSFM} for \ac{MPC}, first in general and then in the context of building energy control.

We consider a finite-horizon \ac{MPC} problem with system outputs $\mathbf y_t \in \mathbb{R}^{n_y}$ and control inputs $\mathbf u_t \in \mathbb{R}^{n_u}$. At time $t$, the controller selects a sequence of future control inputs by optimizing
\begin{mini}|s|
    {\mathbf u_{t:t+N-1}}
    {\mathcal J\!\left(
        \hat{\mathbf y}_{t+1:t+N},
        \mathbf u_{t:t+N-1}
    \right)}
    {\label{eq:mpc}}
    {}
    \addConstraint{
        g\!\left(
            \hat{\mathbf y}_{t+1:t+N},
            \mathbf u_{t:t+N-1}
        \right)
    }{
        \leq 0 ,
    }
\end{mini}
where $\mathcal J: \mathbb R^{N\times n_y}\times\mathbb R^{N\times n_u}\to\mathbb R$ denotes the finite-horizon objective, $g: \mathbb R^{N\times n_y}\times\mathbb R^{N\times n_u}\to\mathbb R^{n_g}$ collects the designed constraints, and $N$ is the prediction horizon.

The output trajectory is predicted by a model
$\mathcal M$ as
\begin{equation}
\hat{\mathbf y}_{t+1:t+N}
=
\mathcal M\!\left(
    \mathbf y_{t-L+1:t},
    \mathbf u_{t-L+1:t+N-1},
    \mathbf d_{t-L+1:t+N}
\right),
\label{eq:predictive_model}
\end{equation}
where $L$ denotes the context length. For compactness, $\mathbf y_{a:b}$ denotes a trajectory slice, and analogously for $\mathbf u_{a:b}$ and $\mathbf d_{a:b}$. The model predicts future outputs from recent measurements $\mathbf y_{t-L+1:t}$, past and candidate future control inputs $\mathbf u_{t-L+1:t}$, and exogenous inputs $\mathbf d_{t-L+1:t+N} \in \mathbb{R}^{(L + N) \times n_d}$ that affect the system but are not controlled. Examples in the \ac{HVAC} setting considered in this paper include outdoor temperature, solar irradiance, and occupancy-related gains. Past quantities are observed, future exogenous inputs are assumed known or forecasted, and $\mathbf u_{t:t+N-1}$ is optimized by the \ac{MPC} planning algorithm.

The quality of the control decision depends on the models ability to correctly predict how the future output trajectory changes under the control sequences considered by the optimizer. This is especially important for designing constraint functions $g$, which often only meaningfully encode operational constraints under an accurate model $\mathcal{M}$.

\subsection{Time-series Foundation Models}

\paragraph{\Acfp{TSFM}.} Pretraining on large corpora of synthetic and real-world data enables \acp{TSFM} to use \ac{ICL}, i.e., to infer a forecasting task from the time-series context without task-specific parameter updates. This enables zero-shot forecasting on unseen univariate and multivariate time-series prediction tasks, as demonstrated by models such as TimeGPT~\citep{garza2023timegpt}, Time-LLM~\citep{jin2024time}, Lag-Llama~\citep{rasul2023lag}, TimesFM~\citep{das2023decoder}, and Chronos~\citep{ansari2024chronos}. Recent approaches differ in how they support multivariate forecasting and known-future covariates: TimesFM~\citep{das2023decoder} is fundamentally univariate, with covariate support provided through an additional regression component; Chronos-2~\citep{chronos2} natively supports multivariate forecasting and known-future covariates; and TabPFN-TS~\citep{hoo2025tables} provides these capabilities through its time-series-to-tabular formulation.

\paragraph{\Acp{TSFM} for \acf{MPC}.} These capabilities are particularly attractive for predictive control, since candidate control trajectories can be provided as known-future covariates. Recent work has explored this connection through action-conditioned \ac{TSFM} dynamics models~\citep{shao2026action}, distillation into compact control models~\citep{liang2026thermostill,park6433759foundation}, and explicit separation of passive dynamics and action effects~\citep{wang2026thermoforce}. However, accurate forecasting of the factual trajectory does not imply accurate prediction under alternative control actions. For \ac{ICL}-based models, this depends critically on whether the inference context contains sufficient variation in the control inputs to identify their effect on the system. In Eq.~\eqref{eq:predictive_model}, $\mathbf y$ therefore corresponds to the prediction targets, while $\mathbf u$ and $\mathbf d$ enter as covariates, with candidate controls and forecast disturbances provided as known-future covariates.

\paragraph{System identification and excitation.} Deploying \ac{MPC} in a dynamical system generally requires first identifying a model from input-output data. The input must excite the system across the frequency range relevant to the deployment conditions and, for nonlinear systems, additionally cover the relevant amplitude range and operating region. The resulting system response must further be distinguishable from the disturbances acting on it. Different excitation strategies address these requirements; \ac{APRBS}, for instance, have been proposed for generating training data for neural-network models of nonlinear systems~\citep{nelles1995identification}. Historical data collected under closed-loop operation is informative only under specific conditions that do not generally hold, motivating deliberate experiment design.

\subsection{Building Energy Management}\label{sec:bldg_en_mgmt}

\paragraph{MPC for building energy management.} \Ac{MPC} is a widely used approach for building energy management, where predictions of thermal and energy dynamics are used to optimize control trajectories subject to comfort and operational constraints. Rather than identifying a state-space model and recursively rolling it out, the multi-step map from past measurements and future inputs to future outputs can be learned directly, as in subspace predictive control~\citep{favoreel1999spc} and related data-driven approaches~\citep{sawant2023model,reinhardt2025data}. We use such a model as a linear baseline:
\begin{equation}
\hat{\mathbf y}_{t+1:t+N}
=
\mathcal A_{\beta_y}^{(y)}
\!\left(\mathbf y_{t-L+1:t}\right)
+
\mathcal A_{\beta_u}^{(u)}
\!\left(\mathbf u_{t-L+1:t+N-1}\right)
+
\mathcal A_{\beta_d}^{(d)}
\!\left(\mathbf d_{t-L+1:t+N}\right)
+
\mathbf \beta_\mathrm{bias},
\label{eq:linear_predictor}
\end{equation}
where $\mathcal A_{\beta_y}^{(y)}$, $\mathcal A_{\beta_u}^{(u)}$, and $\mathcal A_{\beta_d}^{(d)}$ denote learned linear operators acting on the target, control, and exogenous-input trajectories, respectively, and $\beta_\mathrm{bias}$ is a learned bias term. This parameterization directly represents the multi-step response to the control input and therefore provides a natural reference for evaluating whether a \ac{TSFM} captures the same control-relevant relationship.

\paragraph{\Acp{TSFM} for building energy management.} \Acp{TSFM} are attractive for building applications because they can provide accurate forecasts with little or no building-specific training data and have demonstrated strong performance on building-energy prediction tasks~\citep{park2025probabilistic,kreusel2026covariates,han2026unica}. Their use for control is less direct. Prior work has reported difficulties in enforcing physical constraints~\citep{liang2026thermostill} and in reproducing basic control-relevant relationships, such as the response of room temperature to heat-pump supply temperature~\citep{wang2026thermoforce}.

Existing approaches therefore modify how \acp{TSFM} enter the control loop. \citet{liang2026thermostill} distill their predictive knowledge into an RC model that can subsequently be used within \ac{MPC}. \citet{wang2026thermoforce} instead decompose the dynamics into a free component represented by the \ac{TSFM} and a forced component represented by a constrained linear model. The latter is closer to our setting because this method uses the \ac{TSFM} as part of its predictive model, but deliberately avoids relying on it to estimate the response to alternative controls. In contrast, we investigate whether the \ac{TSFM} itself can recover the input-output relationship when supplied with sufficiently informative system-identification data in its inference context.

\begin{figure}[!t]
    \centering
    \includegraphics[width=0.8\linewidth]{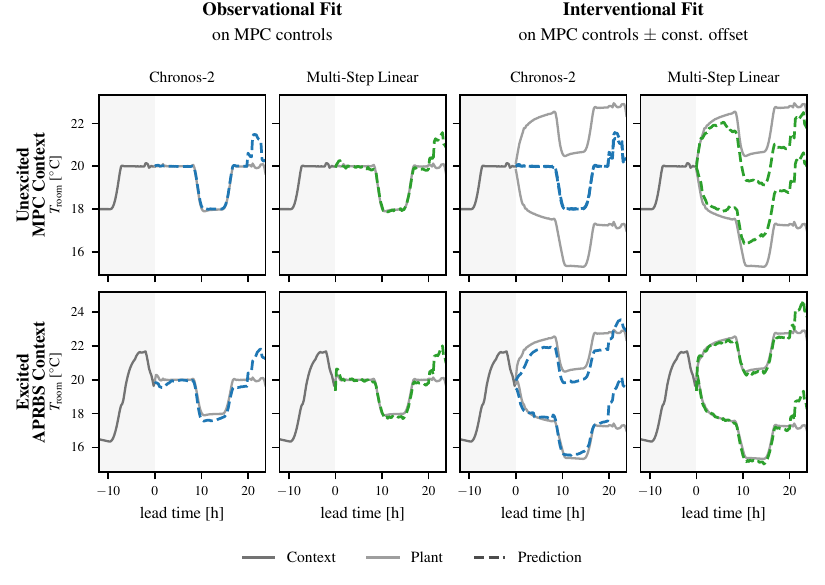}
    \caption{\textbf{Excitation enables Chronos-2 to identify the control effect.}
    Zero-shot Chronos-2 and a multi-step linear estimator predicting $T_\text{room}$ from 21 days of unexcited MPC context (top) or excited APRBS context (bottom).
    Left: observational fit on the MPC control.
    Right: interventional fit, with constant offsets of $±6K$ added to the reference MPC control. Under an unexcited context Chronos-2 fits the observational trajectory well but does not respond to offset-interventions on the controls; the linear estimator, on the same context, does separate them.
    With an excited context Chronos-2 recovers the response.
    The underlying optimization problem allows for a lower comfort temperature bound at night, resulting in periodically lower $T_\text{room}$.
    }
    \label{fig:qualitative}
\end{figure}

\section{Can Time-Series Foundation Models Identify Control Relevant Models?}

In the following, we investigate whether \acfp{TSFM} can identify predictive models that capture the input-output behavior relevant for control.

\subsection{\Acp{TSFM} Can Do  Identification from Excited System Data}

We use the \texttt{i4b}-simulator \cite{zhang2026constrained} to generate heat-pump control time series; states, covariates, and further simulation details are given in \autoref{app:simulation}.
The \ac{TSFM} predicts room temperature $T_\text{room}$, and we evaluate both forecast accuracy and its response to changes in the future control covariate, the heat-pump supply-temperature setpoint $T_\text{hp,supply}$. All projected future disturbances (i.e. weather forecasts) are provided as future covariates. We additionally evaluate a feature-engineered variant with three \ac{EWMA} control channels at different time scales (\autoref{app:simulation}).
Evaluation tasks are based on rollouts of the default \texttt{i4b} \ac{MPC}, which optimizes energy consumption subject to room-temperature comfort bounds. Unlike the multi-step predictor in \autoref{sec:tsfm_mpc}, this controller uses the same recursive one-step dynamics model as the simulator. Throughout the open-loop evaluation, we fix the prediction horizon to 24\,h and evaluate context lengths $L\in\{1,2,5,21\}$ days.

From the reference rollouts, we construct tasks in which the control sequence in the context, forecast window, or both is replaced and the plant re-simulated under the intervention.
For the context, we replace the \ac{MPC} controls with an \ac{APRBS} excitation signal; its amplitudes, dwell-time distribution, and randomization are detailed in \autoref{app:simulation}.
For the forecast window, we use constant offsets to the reference \ac{MPC} controls to evaluate model sensitivity to control inputs.

\begin{figure}[!t]
    \centering
    \includegraphics[width=0.8\linewidth]{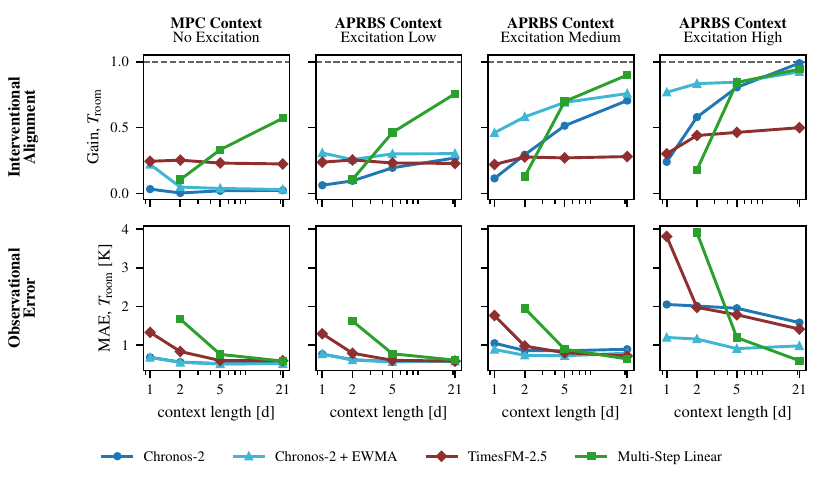}
    \caption{\textbf{In-context identification needs excitation.} Response gain (top; ideal value $1$, dashed) and base-trajectory \ac{MAE} (bottom) against context length, with columns ordered by increasing excitation of the observed context and the known future control held identical throughout. The foundation models recover a control response only under excitation, and only the per-window linear estimator's data requirement, which cannot be fitted on short contexts, keeps them competitive.}
    \label{fig:context-ladder}
\end{figure}

The different settings are illustrated in \autoref{fig:qualitative} where we compare Chronos-2 against a multi-step linear prediction model, fit on the same context window (see \autoref{sec:tsfm_mpc}).
The top row shows the predictions when letting Chronos-2 identify the system from data collected in a \ac{MPC}-rollout while the bottom row shows predictions based on an \ac{APRBS}-generated context.
Under \ac{MPC}-context both Chronos-2 and the multi-step model yield a highly accurate fit of the future trajectory under that same policy (Observational Fit).
However, Chronos-2 does not predict any effect from the control-offset intervention, seemingly unable to identify a correlation between control covariate and target (Interventional Fit).
In contrast, the multi-step linear predictor recovers at least a significant part of the control's effect on the target.
We note that the class of mappings representable by Chronos-2 is much richer, whereas we expect the inductive bias of the multi-step linear model to be very close to the true dynamics of the simulator, making it a very competitive baseline to beat.
With excited context, the ability of Chronos-2 to estimate the effect of the control intervention greatly improves (\autoref{fig:qualitative} bottom row, \ac{APRBS} Context).

\subsection{Measuring Control Fidelity}

Forecasting and control pose different predictive tasks. Forecasting is \emph{observational}: it predicts outcomes under the data-generating policy. Control is \emph{interventional}: it predicts outcomes under alternative inputs. In closed-loop data, the controller selects the supply temperature from the observed state, making their effects difficult to disentangle. Accuracy metrics such as \ac{MAE} under a nominal control sequence therefore do not reveal whether predictions respond correctly when the action changes, see \autoref{fig:qualitative}.

We evaluate this response using sets of $K$ matched \emph{probes} that share the building, context, initial state, and weather but differ in their future control. Each probe is simulated to obtain its ground-truth intervention outcome. After centering the realised and predicted trajectories around their per-step probe means, we compute the \emph{response gain}: the pooled least-squares coefficient of predicted on realised deviations. Gain near $1$ indicates a correctly scaled response, whereas gain near $0$ indicates little sensitivity to the control. \autoref{app:metrics} provides the exact
construction.

Response gain complements rather than replaces nominal forecast error: it removes errors shared across probes and evaluates only the response component. It therefore tests a necessary condition for model-based control, but does not establish complete trajectory agreement or closed-loop control performance.

\subsection{How Much Excitation (Data) is Enough?}

\begin{figure}[!t]
    \centering
    \includegraphics[width=0.8\linewidth]{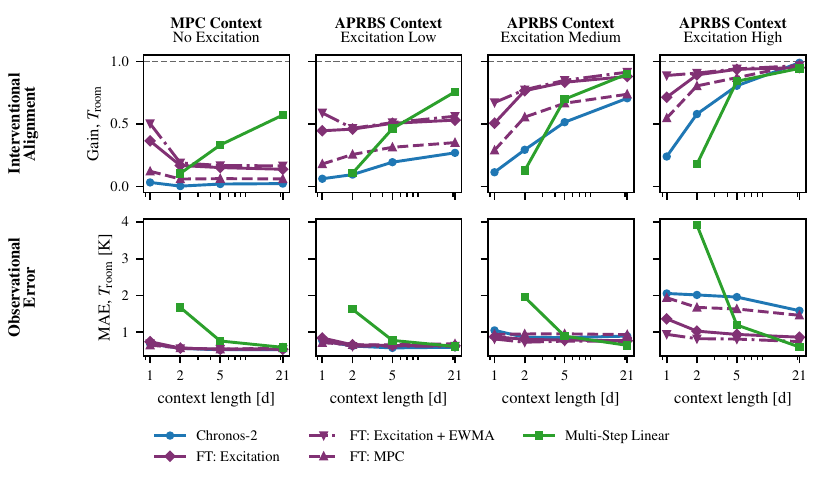}
    \caption{\textbf{Fine-tuning and \ac{EWMA} features reduce, but do not eliminate, the need for in-context excitation.} Response gain (top; ideal value $1$, dashed) and nominal-trajectory \ac{MAE} (bottom) across context lengths, with columns ordered by increasing context excitation. Fine-tuning provides the largest benefit at short context lengths. Combining excited fine-tuning with \ac{EWMA} features recovers substantial gain from one day of context even under no or low excitation, and performs best under medium and high excitation at short-to-intermediate contexts. Under unexcited operation, however, gain remains below one and does not improve with additional context.}
    \label{fig:finetuning}
\end{figure}

Having found that \acp{TSFM} can identify the effect of interventions on control sequences from excitation context we next aim to determine for how long and with what amplitude we must excite the system. One would usually want to minimize both quantities, as they require us to deviate from the primary control objective(s), such as increasing energy cost and reducing thermal comfort.

We find that increasing context length alone without excited data reduces the predictive error of all \acp{TSFM} on the trajectory of the nominal control sequence. But it does not improve their gain (see \autoref{fig:context-ladder}), i.e. their response to changes in the input signal.
Under excitation, gain increases with both context length and excitation amplitude, approaching $1$, or perfect gain, under high excitation and maximum context length. Using \ac{EWMA} features for Chronos-2 significantly reduces the context length it requires to achieve the same gain level. However, it does not increase the maximum gain obtainable at a given excitation level. In our experiments, sufficient excitation amplitude, rather than context length alone, was required for reliable in-context system identification.

This improvement comes at the cost of nominal accuracy: \ac{MAE} increases across excitation regimes, although longer context reduces \ac{MAE} within each regime. The per-window linear estimator achieves the strongest gain over most of the sweep and the lowest long-context \ac{MAE} in every regime, but cannot be fitted reliably with little context. Chronos-2 matches or exceeds its gain only under the strongest excitation with the longest context; elsewhere, the \acp{TSFM} underestimate the response magnitude. Overall, in-context identification requires informative and potentially disruptive excitation and rarely outperforms explicit identification when sufficient data are available. This motivates learning the response once from excited fine-tuning data rather than repeatedly exciting the system at deployment.

\subsection{Can Fine-Tuning Compensate for Limited Excitation?}

\begin{table}[t]
  \centering
  \small
  \setlength{\tabcolsep}{4pt}
  \caption{\textbf{Closed-loop performance relative to the multi-step linear baseline.}
Mean paired differences over $n=17$ held-out buildings under matched evaluation conditions; negative is better. Values are mean $\pm$ standard error. Here, $E$ denotes total heat-pump energy use, and $D$ the time-integrated squared comfort-bound violation. The objective is $\mathrm{Obj.}=E+10D$. Absolute baseline values are shown in the final block.}
  \label{tab:closed_loop_delta}
  \resizebox{0.9\textwidth}{!}{%
  \begin{tabular}{@{}llrrrrrr@{}}
    \toprule
    & & \multicolumn{3}{c}{Initial Context is Unexcited} & \multicolumn{3}{c}{Initial Context is Excited} \\
    \cmidrule(lr){3-5} \cmidrule(lr){6-8}
    Predictor & Ctx & $\Delta E$ [kWh] & $\Delta D$ [K$^2$h] & $\Delta$Obj.
                    & $\Delta E$ [kWh] & $\Delta D$ [K$^2$h] & $\Delta$Obj. \\
    \midrule
    Chronos-2 (zero-shot) & 2\,d & $-4.8 \pm 2.3$ & $-13.6 \pm 6.1$ & $-141.0 \pm 60.7$
                                 & $-6.9 \pm 3.1$ & $-28.4 \pm 12.3$ & $-291.2 \pm 124.4$ \\
                          & 5\,d & $-3.4 \pm 2.2$ & $+14.2 \pm 5.7$ & $+138.7 \pm 58.1$
                                 & $-3.4 \pm 1.6$ & $+6.8 \pm 3.2$ & $+64.1 \pm 32.8$ \\
    \addlinespace
    \quad + EWMA          & 2\,d & $-6.2 \pm 2.2$ & $-21.0 \pm 7.9$ & $-216.3 \pm 78.9$
                                 & $-6.9 \pm 3.1$ & $-34.9 \pm 12.0$ & $-356.0 \pm 120.7$ \\
                          & 5\,d & $-5.4 \pm 1.8$ & $-0.9 \pm 1.2$ & $-14.8 \pm 12.1$
                                 & $-4.2 \pm 1.6$ & $-0.0 \pm 0.7$ & $-4.3 \pm 7.3$ \\
    \addlinespace
    Chronos-2 (fine-tuned)            & 2\,d & $-4.7 \pm 2.3$ & $-18.9 \pm 7.0$ & $-193.4 \pm 70.6$
                                 & $-6.9 \pm 3.0$ & $-32.2 \pm 12.0$ & $-328.6 \pm 120.9$ \\
                          & 5\,d & $-4.3 \pm 1.5$ & $+0.0 \pm 1.6$ & $-4.2 \pm 16.3$
                                 & $-4.9 \pm 1.7$ & $-1.2 \pm 0.7$ & $\mathbf{-16.8 \pm 7.5}$ \\
    \addlinespace
    \quad + EWMA          & 2\,d & $-5.8 \pm 2.1$ & $-21.8 \pm 8.1$ & $-223.8 \pm 81.6$
                                 & $-7.0 \pm 3.0$ & $-35.5 \pm 12.0$ & $\mathbf{-362.2 \pm 121.0}$ \\
                          & 5\,d & $-4.2 \pm 1.6$ & $-2.9 \pm 1.4$ & $\mathbf{-33.0 \pm 14.5}$
                                 & $-4.0 \pm 1.7$ & $-1.2 \pm 0.8$ & $-15.8 \pm 8.2$ \\
    \midrule
    Multi-Step Linear     & 2\,d & $25.4$ & $33.9$ & $364.8$ & $27.7$ & $47.3$ & $500.2$ \\
    \emph{(absolute)}     & 5\,d & $43.6$ & $13.5$ & $178.8$ & $43.1$ & $11.9$ & $162.2$ \\
    \bottomrule
  \end{tabular}
  }
\end{table}

The previous results show that in-context identification can require disruptive excitation at deployment. We therefore test whether cross-building fine-tuning can reduce this requirement. \autoref{fig:finetuning} compares zero-shot Chronos-2 with models fine-tuned on closed-loop \ac{MPC} or excited data. Fine-tuning improves response gain primarily when context is limited, with excited fine-tuning outperforming \ac{MPC}-only fine-tuning. Adding \ac{EWMA} features yields the strongest gains: with one day of context, this variant already recovers a substantial response under no or low excitation, and under medium or high excitation it dominates at short-to-intermediate context lengths. These improvements incur no systematic accuracy penalty, as its nominal \ac{MAE} is generally lowest or close to lowest.

Fine-tuning nevertheless does not fully replace excitation. Under unexcited operation, gain remains well below $1$ and does not improve with additional context; given sufficient history, the per-window linear estimator eventually performs better. By contrast, medium or high excitation allows the fine-tuned \ac{EWMA} model to approach unit gain, with differences between models diminishing as context grows. Fine-tuning and multiscale features therefore transfer useful prior information and substantially reduce the required context and excitation, but informative variation in the target system remains necessary for full identification.

\subsection{Preliminary Closed-Loop Evaluation}

As a preliminary and separate closed-loop evaluation, we compare the predictors within \ac{MPC} on held-out buildings. We use \ac{iCEM}~\citep{icem} to optimize the control sequence and compare all predictors against the multi-step linear baseline under matched evaluation conditions. \autoref{tab:closed_loop_delta} shows that Chronos-2 can improve the closed-loop objective, particularly with an excited context and \ac{EWMA} correction. These experiments do not establish a direct relationship between open-loop response gain and closed-loop performance.

Surprisingly, even when the initial context at the start of an episode is unexcited, the performance is good. We hypothesize that this arises because future control actions selected by \ac{iCEM} are subsequently executed and enter the rolling context, thereby providing implicit excitation. In contrast, the data used as \ac{MPC} context in the open-loop experiments is generated by a deterministic \ac{MPC}~\cite{zhang2026constrained}, with no learning or inference dynamics. Verifying this mechanism directly is left for future work. Full experimental details and additional results are provided in \autoref{app:simulation}.

\section{Conclusion}

We studied whether \acp{TSFM} can serve as predictive models for \ac{MPC}, using building thermal control as a test bed. Our results show that \acp{TSFM} can recover control-relevant input-output relationships when their context contains sufficient excitation, whereas nominal forecast accuracy alone does not reveal this ability. Fine-tuning on excited trajectories and multiscale \ac{EWMA} control features substantially improve open-loop input-response fidelity and closed-loop control performance, while reducing the amount of informative context required. Nevertheless, reliable identification still depends on informative variation in the target system, and a compact linear identifier remains competitive given sufficient excited history.

Our evaluation is primarily synthetic, leaving open how well these findings transfer to real-world systems. Future work should investigate \acp{TSFM} designed specifically for control, including whether naturally occurring control variation, such as price-responsive operation, and fine-tuning can further reduce the need for dedicated excitation. More broadly, our results suggest that context is a central design variable for \ac{TSFM}-based control: how to retain control-informative observations and how much excitation is required for reliable
input-output identification remain open questions.

\section*{Acknowledgment}

Mazen Amria was supported by the German Academic Exchange Service (DAAD) through the ``Study Scholarships~\textendash~Master Studies for All Academic Disciplines, 2024/25'' programme (Funding Programme No. 57693457). This work was conducted during the period of support provided by this scholarship.
The authors further acknowledge the Institute for Machine-Brain Interfacing Technology (IMBIT), University of Freiburg, for providing research infrastructure. Computational resources: Gefördert durch die Deutsche Forschungsgemeinschaft (DFG) – 428605208. 

\bibliographystyle{unsrtnat}
\bibliography{references}


\clearpage
\appendix

\section{Real-World Data \& Real-World Probing}
\label{app:real_probe}

\paragraph{Real-world data.} In this work, we evaluate on measurements from a single-family house in Trondheim, Norway, equipped with a ground-source heat pump. The record spans 15 April to 24 August 2026 --- 131 days at a 20-minute cadence, 9{,}468 samples --- and carries indoor and outdoor temperature, the heat pump's supply and return temperatures, the requested supply setpoint, global irradiance and electrical power. The house is driven throughout by its own controller, so the data contain no excitation and no intervention specifically for the context of this work. 

\begin{figure}[!htbp]
    \centering
    \includegraphics[width=0.95\textwidth]{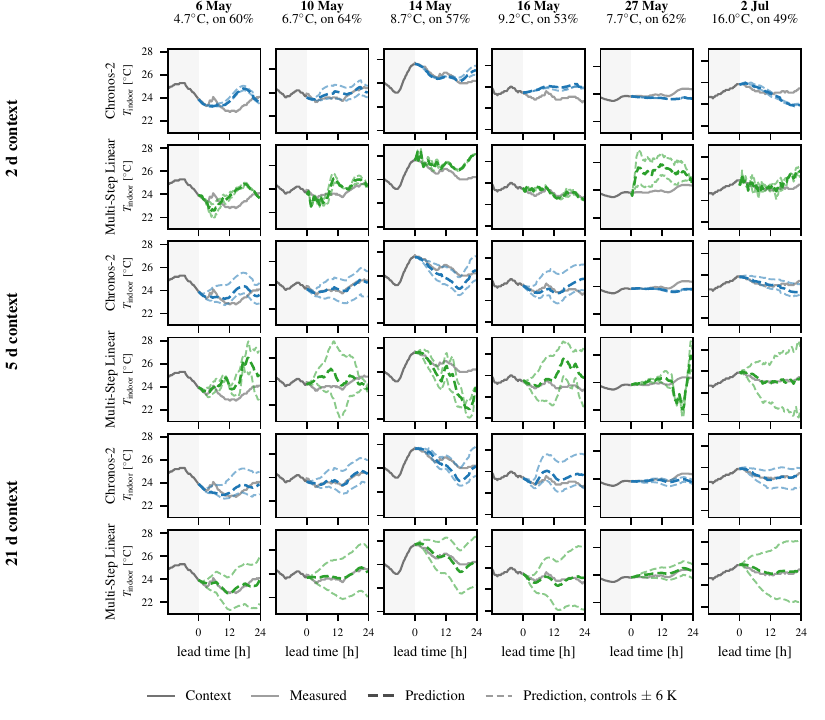}
    \caption{\textbf{Control response on a real building.} Indoor temperature forecast 24\,h ahead
    under the measured supply setpoint (heavy dashed) and under that setpoint shifted by
    $\pm 6$\,K (light dashed), against the measurement (grey). Columns are the six most
    heating-active days at least 48\,h apart, headed by their mean outdoor temperature and the
    share of the horizon the heat pump spent heating; rows cross context length with predictor,
    and each column shares a $y$ axis. With a short context both predictors are almost insensitive
    to the control. The house ran under its own controller, so the offset plans have no
    measured response to score against and the figure shows only whether the forecast moves. The percentage under the date describes the time the heat-pump was active in the context.}
    \label{fig:intervention_real}
\end{figure}

\paragraph{Does a response appear on a real building?}
\autoref{fig:intervention_real} tests different control sequences on the measured house, forecasting
indoor temperature under the measured supply setpoint and under that setpoint shifted by
$\pm 6$\,K, at three context lengths. The house ran once, so the offset plans have no measured
response to score against and the figure asks only whether the forecast moves when the plan moves.
The broad trend matches the simulated results: with a short context both predictors are almost
insensitive to the control, and only with longer context they respond.

\vspace{1cm}

\clearpage

\section{Metrics Details \& Control Interventions}
\label{app:metrics}

In the following, we provide detailed information about how we did control interventions and defined metrics used throughout the paper.

Throughout, $W$ indexes evaluation windows, $H = 96$ the horizon steps and $K = 7$ the probes per
window. We denote for probe $k$ with $u^{(k)}$ the control trajectory, $y^{(k)}$ the plant's realised room temperature
$T_\text{room}$ and $\hat{y}^{(k)}$ the model's prediction. The index $n$ marks the
\emph{nominal} probe, the control the data corpus actually applied. All metrics are reported only based on $T_\text{room}$.

\paragraph{Probe construction.} From a single anchor window, and holding the context and the
weather fixed, the probes differ only in the control. Five are constant offsets,
\begin{equation*}
    u^{(k)} \;=\; \mathrm{clip}\!\left(u^{(n)} + \delta_k,\; 5\,^\circ\mathrm{C},\; 65\,^\circ\mathrm{C}\right),
    \qquad \delta_k \in \{-6, -2, 0, 2, 6\}\ \mathrm{K},
\end{equation*}
and two replace the constant by an \ac{APRBS} of the same $6$\,K amplitude, added and subtracted,
so that the probe set separates a model that tracks the control's shape from one that merely
absorbs a constant shift. The clip enforces the physical actuator range of the heat pump. Each
$u^{(k)}$ is rolled through the simulator, so $y^{(k)}$ is the trajectory the building would
really have followed under that control, making the probes interventions on the plant rather than
counterfactuals inferred from a model.

\paragraph{Accuracy.}
MAE is computed on the nominal probe only, averaged over the horizon and then over windows:
\begin{equation}
  \text{MAE} \;=\; \frac{1}{W}\sum_{w=1}^{W}\frac{1}{H}\sum_{t=1}^{H}
    \bigl|\hat{y}^{(n)}_{w,t} - y^{(n)}_{w,t}\bigr|.
\end{equation}
It therefore measures the fit to the trajectory the plant actually followed, and carries no
information about how the model responds to a change of plan.

\paragraph{Control response.}
Deviations are taken about the probe mean at each step, separately for the plant and the model:
\begin{equation}
  \Delta y^{(k)}_{w,t} \;=\; y^{(k)}_{w,t} - \frac{1}{K}\sum_{j=1}^{K} y^{(j)}_{w,t} ,
  \qquad
  \Delta\hat{y}^{(k)}_{w,t} \;=\; \hat{y}^{(k)}_{w,t} - \frac{1}{K}\sum_{j=1}^{K} \hat{y}^{(j)}_{w,t} .
  \label{eq:deviations}
\end{equation}
Because the probes share their context and disturbances, this subtraction removes components shared
across probes, including any probe-invariant offset in the model's predictions, and isolates
variation associated with the control interventions. Response gain is the pooled least-squares
coefficient of the predicted deviation on the realised one,
\begin{equation}
  \mathrm{Gain} \;=\;
  \frac{\sum_{w,k,t} \Delta\hat{y}^{(k)}_{w,t}\,\Delta y^{(k)}_{w,t}}
       {\sum_{w,k,t} \bigl(\Delta y^{(k)}_{w,t}\bigr)^{2}} .
  \label{eq:gain}
\end{equation}
The sums run over windows as well as probes and horizon steps, so Eq.~\eqref{eq:gain} is one pooled
coefficient rather than an average of per-window ratios: windows in which the plant responds
strongly carry more weight, and a window in which the control did little cannot contribute an
unstable ratio.

\paragraph{Interpreting the gain.}
A value of $1$ indicates correct scaling along the simulated response direction. It does not rule
out residual response error, because gain measures only the component of the prediction aligned
with the plant's. A value of $0$ indicates no aligned response, but does not imply that the
prediction is insensitive to $u$ --- a prediction that responded orthogonally to the plant would
also score zero. Values above $1$ indicate over-response rather than further improvement. Because
Eq.~\eqref{eq:deviations} removes everything common to the probes, gain is insensitive to error shared
across them: a model cannot score well on it by being accurate, and cannot be penalised on it for
a constant bias. Nominal forecast error and response gain therefore measure distinct aspects of
predictive performance and should be read together.

\clearpage

\section{Synthetic Setup}\label{app:simulation}

This section describes the synthetic building population, weather data, model inputs, fine-tuning procedure, and open- and closed-loop evaluation setup.

\paragraph{Building population and data.}
We simulate 191 single-family buildings in \texttt{i4b} using parameters from
the TABULA database \cite{zhang2026constrained}. The population covers
seven European countries, construction years from the 1910s to the 2010s, and
nearly an order of magnitude in transmission loss. Each building is represented
by a 4R3C thermal model and coupled to an air-source heat pump with a water-based
emitter circuit. The controlled input is the heat-pump supply-temperature
setpoint $T_\mathrm{hp,sup}$, while modeled outputs are room temperature
$T_\mathrm{room}$ and heat-pump return temperature $T_\mathrm{hp,ret}$.

The population is split into 131 training, 30 validation, and 30 test buildings,
yielding 1{,}441, 330, and 330 trajectories, respectively. No building appears
in more than one split. Training and validation use April 2024--March 2025,
while testing uses April 2025--March 2026, so test buildings are unseen in both
building parameters and weather year. All simulations use a $15$-minute
sampling interval.

\paragraph{Weather data.}
For each of the seven locations, we use ambient temperature, global horizontal
irradiance (GHI), direct normal irradiance (DNI), and diffuse horizontal
irradiance (DHI). Realized weather is obtained from the ERA5-backed Open-Meteo
archive, while archived forecasts use ECMWF IFS runs initialized at 00 and
12 UTC with a $48$\,h horizon. A $6$\,h availability delay ensures that only
forecasts available at the corresponding decision time are exposed to the
controller.

\begin{table}[H]
  \centering
  \small
  \caption{TSFM channels used throughout the experiments.}
  \label{tab:channels}
  \begin{tabular}{@{}lllll@{}}
    \toprule
    \textbf{Channel} & \textbf{TSFM type} & \textbf{Control role}
    & \textbf{Context} & \textbf{Horizon} \\
    \midrule
    $T_\mathrm{room}$               & target    & output    & observed & predicted \\
    $T_\mathrm{hp,ret}$             & target    & output    & observed & predicted \\
    $T_\mathrm{hp,sup}$             & covariate & control   & observed & given \\
    $T_\mathrm{amb}$, GHI, DNI, DHI & covariate & exogenous & measured & forecast \\
    \bottomrule
  \end{tabular}
\end{table}

\autoref{tab:channels} summarizes the model channels. In particular,
$T_\mathrm{hp,sup}$ is supplied over the prediction horizon as the candidate
control trajectory.

\paragraph{Training data and fine-tuning.}
Training trajectories are generated under nominal and perturbed MPC operation
and open-loop APRBS excitation. We sample 31 windows from each of the 131
training buildings, yielding 4{,}061 items. Each item contains up to $21$\,d of
context and a $24$\,h prediction horizon, with the context length sampled
uniformly between $1$ and $21$ days. Fine-tuned variants adapt Chronos-2 with
LoRA while keeping the base weights frozen. We train for $2{,}000$ steps with
batch size 256 and learning rate $10^{-5}$, supervising
$T_\mathrm{room}$ and $T_\mathrm{hp,ret}$.

\paragraph{Multiscale control features.}
The \acf{EWMA} variant augments $T_\mathrm{hp,sup}$ with three causal moving
averages,
\begin{equation*}
  u^{(s)}_t
  =
  \alpha_s u_t + (1-\alpha_s)u^{(s)}_{t-1},
  \qquad
  \alpha_s = \frac{2}{s+1},
  \qquad
  s \in \{2,8,32\},
\end{equation*}
corresponding to $30$\,min, $2$\,h, and $8$\,h. These features expose the
accumulated control input over multiple time scales and can be evaluated
causally along candidate future control trajectories.

\paragraph{Open-loop evaluation.}
We evaluate input-response fidelity on 240 cases sampled from the held-out test
corpus, using a $24$\,h horizon and context lengths of $1$, $2$, $5$, and
$21$ days. For each case, we start from the same nominal MPC control trajectory
$\mathbf u^{n}$ and construct six matched probes as
\begin{equation}
    \mathbf u^{(k,\pm)}
    =
    \mathbf u^{(n)}
    \pm a_k \mathbf q_k,
    \qquad
    (a_k,\mathbf q_k)
    \in
    \left\{
        (2\,\mathrm{K},\mathbf 1),
        (6\,\mathrm{K},\mathbf 1),
        (6\,\mathrm{K},\mathbf q_\mathrm{APRBS})
    \right\}.
    \label{eq:control_probes}
\end{equation}
Together with the nominal trajectory, this gives seven probes per case. All
probes share the same initial condition and exogenous inputs, so their response
differences isolate the effect of the candidate control trajectory. We evaluate
nominal forecast error and the agreement between predicted and simulated
$T_\mathrm{room}$ responses to these interventions.

\paragraph{Excitation levels in context.}
We compare contexts generated under nominal MPC with contexts generated under
pure \ac{APRBS} excitation. In the latter, the heat-pump supply-temperature
setpoint itself follows a piecewise-constant \ac{APRBS} signal, with dwell times
sampled uniformly between $1$ and $6$\,h. We use excitation amplitudes of
$1.5$, $6$, and $24$\,K for the low, medium, and high used throughout this work. For each building and period, the three levels use
the same \ac{APRBS} realization scaled in amplitude, so the comparison isolates the
amount of excitation present in the context.

\paragraph{Closed-loop evaluation.}
Closed-loop control is evaluated on 17 test buildings using two-day
episodes, a $24$\,h planning horizon, and replanning every $15$ minutes.
We use rolling contexts of $2$ and $5$ days, initialized from either nominal
operation or an excited trajectory. In both cases, the context is subsequently updated with the observations generated by the applied control actions.

At each step, candidate supply-temperature trajectories are optimized with \ac{iCEM}. The planner uses four iterations, eight elites, and populations of $32$, $26$, $21$, and $16$ trajectories. Candidate trajectories are sampled with temporally correlated $1/f^{2.5}$ noise. The search distribution and elites are warm-started between replanning steps. Controls are bounded to $20$--$60\,^\circ$C and further restricted to within $5$\,K of the reference-control range. Candidate plans are ranked lexicographically by comfort violation and energy consumption.

\paragraph{Compute Statement.} All experiments run on nodes with eight NVIDIA RTX A6000 GPUs (48\,GB) and
two AMD EPYC 7542 32-core processors. Fine-tuning takes about $36$ minutes on a single GPU.
Closed-loop evaluation dominates the budget, since the planner scores roughly $95$ candidate
plans at every $15$-minute control step: a two-day episode costs $150$ to $600$\,s per building.


\end{document}